\documentclass{llncs}
\usepackage{url}
\usepackage{graphicx}
\usepackage{booktabs}

\usepackage{hyperref}
\hypersetup{
  bookmarks=true,
  pdftitle={Theorem Proving in Large Formal Mathematics as an Emerging AI Field},
  pdfkeywords={automated theorem proving, proof discovery, proof analysis, formal mathematics, large-theory automated reasoning},
  pdfauthor={Josef Urban, Ji\v r\'{\i} Vysko\v cil},
  pdfsubject={Submitted to LPAR-18},
  colorlinks=true,
  linkcolor=black,
  citecolor=black,
}

\begin{document}
\title{Theorem Proving in Large Formal Mathematics as an Emerging AI Field}
\author{Josef Urban\inst{1}\thanks{Supported by The Netherlands Organization for
Scientific Research (NWO) grants \textit{Knowledge-based Automated Reasoning} and
\textit{MathWiki}.} 
\and Ji\v r\'{\i} Vysko\v cil\inst{2}
\thanks{Supported by the Czech institutional grant MSM 6840770038.}
}
\authorrunning{Urban and Vysko\v cil}

\institute{
Radboud University Nijmegen, The Netherlands
\and
Czech Technical University
}

\tocauthor{Josef Urban, Ji\v r\'{\i} Vysko\v cil}

\maketitle

\begin{abstract}
  In the recent years, we have linked a large corpus of formal
  mathematics with automated theorem proving (ATP) tools, and started to develop combined AI/ATP
  systems working in this setting.  In this paper we first relate this
  project to the earlier large-scale automated developments done by
  Quaife with McCune's Otter system, and to the discussions of the
  QED project about formalizing a significant part of mathematics.
  Then we summarize our adventure so far, argue that the QED dreams were
  right in anticipating the creation of a very interesting semantic AI
  field, and discuss its further research directions.
\end{abstract}

\section{OTTER and QED}
Twenty years ago, in 1992, Art Quaife's book \emph{Automated
  Development of Fundamental Mathematical Theories}~\cite{Qua92-Book} was published. In the conclusion to his JAR paper~\cite{Qua92-JAR} about the development of set theory Quaife cites Hilbert's \emph{``No one shall be able to drive us from the paradise that Cantor created for us''}, and makes the following bold prediction:

\begin{quote}
The time will come when such crushers as Riemann's hypothesis and
Goldbach's conjecture will be fair game for automated reasoning
programs. For those of us who arrange to stick around, endless fun
awaits us in the automated development and eventual enrichment of the
corpus of mathematics.
\end{quote}

Quaife's experiments were done using an ATP system that has left so
far perhaps the greatest influence on the field of Automated Theorem
Proving: Bill McCune's Otter. Bill McCune's opinion on using Otter
and similar Automated Reasoning methods for general mathematics was
much more sober. The Otter manual~\cite{McC03-Otter} (right before acknowledging Quaife's work) states:

\begin{quote}
Some of the first applications that come to mind when one hears ``automated theorem proving'' 
are number theory, calculus, and plane geometry, because these are some of the first
areas in which math students try to prove theorems. Unfortunately, OTTER cannot do much
in these areas: interesting number theory problems usually require induction, interesting
calculus and analysis problems usually require higher-order functions, and the first-order
axiomatizations of geometry are not practical.
\end{quote}

Yet, Bill McCune was also a part of the QED\footnote{\url{http://www.rbjones.com/rbjpub/logic/qedres00.htm}}
discussions about making a significant part of
mathematics computer understandable, verified, and available for a
number of applications. These discussions started around 1993, and gained considerable 
attention of the theorem proving and formalization communities,
leading to the publication of the anonymous QED manifesto~\cite{Anonymous94}  and two QED workshops in 1994 and 1995.
And ATP systems such as SPASS~\cite{WeidenbachDFKSW09}, E~\cite{Sch02-AICOMM}, and Vampire~\cite{RV02-AICOMM} based on the ideas that were
first developed in Otter have now been really used for several years to prove
lemmas in general mathematical developments in the large
ATP-translated libraries of Mizar~\cite{mizar-in-a-nutshell} and Isabelle~\cite{NipkowPW02}.

This paper summarizes our experience so far with the QED-inspired
project of developing automated reasoning methods for large general
computer understandable mathematics, particularly in the large Mizar
Mathematical Library.~\footnote{\url{www.mizar.org}}  A bit as Art Quaife did, we believe (and try
to argue below) that automated reasoning in general mathematics is one
of the most exciting research fields, where a number of new and
interesting topics for general AI research emerge today.
We hope that the paper might be of some
interest %
to those QED-dreamers 
who remember the great minds of the
recently deceased Bill McCune, John McCarthy, and N.G. de Bruijn.

\section{Why Link Large Formal Mathematics with AI/ATP Methods?}
Formalization of Mathematics and Automated Theorem Proving are often
considered to be two subfields of the field of Automated Reasoning. In
the former, the focus is on developing methods for (human-assisted)
computer verification of more and more complex parts of mathematics, while in
ATP, the focus is on developing stronger and stronger methods for
finding proofs automatically.
The QED Manifesto has the following conservative opinion about the
usefulness of automated theorem proving (and other AI) methods to a QED-like project:
\begin{quote}
  It is the view of some of us that many people who could have easily
  contributed to project QED have been distracted away by the enticing
  lure of AI or AR. It can be agreed that the grand visions of AI or
  AR are much more interesting than a completed QED system while still
  believing that there is great aesthetic, philosophical, scientific,
  educational, and technological value in the construction of the QED
  system, regardless of whether its construction is or is not largely
  done `by hand' or largely automatically.
\end{quote}

Our opinion is that formalization and automation are two sides of the same coin.
There are three kinds of benefits in linking formal proof assistants like Mizar and their libraries with 
the Automated Reasoning technology and particularly ATPs:

\paragraph{1. The obvious benefits for the proof assistants and their libraries.} Automated Reasoning and AI methods
can provide a number of tools and strong methods that can assist the formalization, provide advanced search and hint functions, and prove lemmas and theorems (semi-)automatically. The QED Manifesto says:
\begin{quote}
The QED system we imagine will provide a means by which mathematicians and scientists can scan the entirety of mathematical knowledge for relevant results and, using tools of the QED system, build upon such results with reliability and confidence but without the need for minute comprehension of the details or even the ultimate foundations of the parts of the system upon which they build
\end{quote}

\paragraph{2. The (a bit less obvious) benefits for the field of Automated Reasoning.} 
Research in automated reasoning over very large formal libraries is painfully theoretical
(and practically useless)
until such libraries are really available for experiments. Mathematicians (and scientists, and other
human ``reasoners'') typically know a lot of things about the domains of discourse, and
use the knowledge in many ways that include many heuristic methods. It thus
seems unrealistic (and limiting) to develop the automated reasoning tools solely for problems that
contain only a few axioms, make little use of previously accumulated knowledge, and do not
attempt to further accumulate and organize the body of knowledge.
In his 1996 review of Quaife's book,
Desmond Fearnley-Sander says:
\begin{quote}
The real work in
proving a deep theorem lies in the development of the theory that it belongs to and its
relationships to other theories, the design of definitions and axioms, the selection of good
inference rules, and the recognition and proof of more basic theorems.
Currently, no resolution-based program, when faced with the stark problem of proving a
hard theorem, can do all this. That is not surprising. No person can either. Remarks about
standing on the shoulders of giants are not just false modesty...
\end{quote}

\paragraph{3. The benefits for the field of general Artificial Intelligence.} These benefits are perhaps
the least mentioned ones, perhaps due to the frequent feeling of too many
unfulfilled promises from general AI.\
However to the authors they appear to be the strongest long-term
motivation for this kind of work. In short, the AI fields of \emph{deductive reasoning} 
and \emph{inductive reasoning} 
(represented by machine learning, data mining, knowledge discovery in databases, etc.)
have 
so far benefited relatively little from each other's progress. This is an obvious deficiency
in comparison with the human mind, which can both inductively suggest new ideas and problem solutions
based on analogy, memory, statistical evidence, etc.,
and also confirm, adjust, and even significantly modify these ideas and problem solutions by deductive 
reasoning and explanation, based on the understanding of the world. 
Repositories of ``human thought''  that are both
large (and thus allow inductive methods), and have precise and deep semantics (and thus allow
deduction) should be a very useful component for cross-fertilization of these two fields. 
QED-like large formal mathematical libraries are currently the closest approximation to such a
computer-understandable repository of ``human thought'' usable for these purposes.
To be really usable, the libraries however again have to be presented
in a form that is easy to understand for existing automated reasoning tools.
The Fearnley-Sander's quote started above continues as:
\begin{quote}
... Great theorems require great theories and theories do not, it seems, emerge from thin air. 
Their creation requires sweat, knowledge, imagination, genius, collaboration and time. As yet there is not much serious collaboration of machines with one another, and we are only just beginning to see real symbiosis between people and machines in the exercise of rationality.
\end{quote}

\section{Why Mizar?}
Mizar is a language for formalization of mathematics and a proof checker for the language, developed since 1974~\cite{mizar-first-30}
by the Mizar team led by Andrzej Trybulec.
This system
was chosen by the first author for experiments with automated reasoning tools
because of its focus on building the large formal  Mizar Mathematical Library (MML).
This formalization effort was begun in 1989 by the Mizar team, and its main purpose
is to verify a large body of mainstream mathematics in a way that is close
and easily understandable to mathematicians, allowing them to build on this library
with proofs from more and more advanced mathematical fields. In 2012 the library consisted of over 1100 formal articles containg more than 50000 proved theorems and 10000 definitions.
The QED discussions often used Mizar and its library as a prototypical example for the project.  
The particular
formalization goals have influenced:
\begin{itemize}
\item the choice of a relatively human-oriented formal language in Mizar
\item the choice of the declarative Mizar proof style (Jaskowski-style natural deduction~\cite{Jaskowski34}) 
\item the choice of first-order logic and set theory as unified common foundations for the whole library
\item  the focus on developing and using just
one human-obvious first-order justification rule in Mizar
\item and the focus on making the large
library interconnected, usable for more advanced formalizations, and using
consistent notation.
\end{itemize}
There have always been other systems and projects that are similar to Mizar in some of the
above mentioned aspects. 
Building large and advanced formal libraries seems to be more and more common today in systems like Isabelle, Coq~\cite{BC04}, and HOL Light~\cite{Harrison96}. 
An example of a recent major formalization project (in HOL Light) is Flyspeck~\cite{Hales05}, which required the formal verification of thousands
of lemmas in general mathematics.
In the work that is described here, Mizar thus should be considered as a suitable particular 
choice of a system
for formalization of mathematics which uses relatively common and accessible foundations, and
produces a large formal library written in a relatively simple and easy-to-understand style.
Some of the systems described below actually already work 
also with other than Mizar data: for example, 
MaLARea~\cite{Urb07-ESARLT,US+08} has already been 
successfully used for reasoning over problems from the large formal SUMO ontology~\cite{NilesP01}, and for experiments with Isabelle/Sledgehammer problems~\cite{MengP09}.

\section{MPTP: Translating Mizar for Automated Reasoning tools}
\label{Translation1}
The Mizar's translation (MPTP - Mizar Problems for Theorem Proving) to pure first-order logic is described in detail
in~\cite{Urb03,Urb04-MPTP0,Urb07}. 
The translation
has to deal with a number of Mizar extensions and practical
issues related to the Mizar implementation, implementations of first-order ATP systems,
and the most frequent uses of the translation system. 

\label{Mptp01}
The first version (published in early 2003\footnote{\url{http://mizar.uwb.edu.pl/forum/archive/0303/msg00004.html}}) has been used 
for initial  exploration of the usability of ATP systems on the Mizar 
Mathematical Library (MML).
The first important number obtained was the 41\% success rate 
of ATP re-proving of about 30000 MML theorems from selected
Mizar theorems and definitions (precisely: other Mizar theorems and definitions mentioned in the 
Mizar proofs)
taken from corresponding MML proofs.

No previous evidence about the feasibility and usefulness of ATP methods on a very large
library like MML was available prior to the experiments done with 
MPTP 0.1\footnote{A lot of work on MPTP has been inspired by the previous work done in the 
ILF project~\cite{Dah97} on importing Mizar. However it seemed that the project
had stopped before it could finish the export of the whole MML to ATP problems and provide
some initial overall statistics of ATP success rate on MML.}, 
sometimes leading to overly pessimistic views on such a project.
Therefore the goal
of this first version was to relatively quickly achieve a ``mostly-correctly'' 
translated version of the whole
MML that would allow us to measure and assess the potential of ATP methods for this large library.
Many shortcuts and simplifications were therefore taken in this first
MPTP version, for example direct encoding in the DFG \cite{HKW96} 
  syntax used by the SPASS system, no proof export,
incomplete export of some relatively rare constructs ({\it structure} types and {\it abstract terms}), etc.

Many of these simplifications however made further experiments with
MPTP difficult or impossible, and also made the 41\% success rate uncertain. 
The lack of proof structure prevented
measurements of ATP success rate on all internal proof lemmas, and 
experiments with unfolding lemmas with their own proofs. 
Additionally, even if only several abstract terms were translated 
incorrectly, during such proof unfoldings they could spread much wider.
Experiments like finding new proofs, and cross-verification of Mizar
proofs (described below) would suffer from constant doubt about
the possible amount of error caused by the incorrectly translated parts of Mizar,
and debugging would be very hard.

Therefore, after the encouraging initial experiments, a new version of
MPTP started to be developed in 2005, requiring first a substantial
re-implementation~\cite{Urb06-mkm} of Mizar interfaces.
This version consists of two layers (Mizar-extended TPTP format
processed in Prolog, and a Mizar XML format) that are sufficiently
flexible and have allowed a number of gradual additions of various
functions over the past years. The experiments described below are typically done on this version (and its extensions).
\label{Mptp02}

\section{Experiments and projects based on the MPTP}
\label{Projects}
MPTP has so far been used for 
\begin{itemize}
\item experiments with re-proving Mizar theorems and simple lemmas by ATPs from
      the theorems and definitions used in the corresponding Mizar proofs,
\item experiments with automated proving of Mizar theorems with the whole Mizar library, i.e. the 
necessary axioms are selected automatically from MML,
\item finding new ATP proofs that are simpler than the original Mizar proofs,
\item ATP-based cross-verification of the Mizar proofs,
\item ATP-based explanation of Mizar atomic inferences,
\item inclusion of Mizar problems in the TPTP~\cite{Sutcliffe10} problem library, and unified web presentation
      of Mizar together with the corresponding TPTP problems,
\item creation of the MPTP \$100 Challenges\footnote{\url{www.tptp.org/MPTPChallenge}} for reasoning in large theories in 2006,  
      creation of the MZR category of the CASC Large Theory Batch (LTB) competition~\cite{Sutcliffe09}
      in 2008, and creation of the MPTP2078~\cite{abs-1108-3446} benchmark\footnote{\url{http://wiki.mizar.org/twiki/bin/view/Mizar/MpTP2078}} in 2011 used for the CASC@Turing\footnote{\url{http://www.cs.miami.edu/~tptp/CASC/J6/Design.html}} competition in 2012,
\item testbed for AI systems like MaLARea and MaLeCoP~\cite{UrbanVS11} targeted at reasoning in large theories and combining
inductive techniques like machine learning with deductive reasoning.
\end{itemize}
\subsection{Re-proving experiments}
As mentioned in Section~\ref{Mptp01}, the initial large-scale experiment done with
MPTP 0.1 indicated that 41\% of the Mizar proofs can be automatically found
by ATPs, if the users provide as axioms to the ATPs the same theorems and definitions
which are used in the Mizar proofs, plus the corresponding background formulas (formulas implicitly used by Mizar, for example to implement type hierarchies). As already 
mentioned, this number was far from certain, e.g., out of the 27449 problems tried, 625
were shown to be counter-satisfiable (i.e., a model of the negated
conjecture and axioms exists) in a relatively low time limit given to SPASS (pointing
to various oversimplifications taken in MPTP 0.1). This experiment was therefore repeated
with MPTP 0.2, however only with 12529 problems that come from articles that do not use
internal arithmetical evaluations done by Mizar. These evaluations were not handled by MPTP
0.2 at the time of conducting these experiments, being the last (known) part of Mizar
that could be blamed for possible ATP incompleteness.
The E prover version 0.9 and SPASS version 2.1 were used for this experiment, with 20s
 time limit (due to limited resources).
The results (reported in \cite{Urb07}) are given in Table \ref{Repr1}.
39\% of the 12529 theorems were proved by either SPASS or E,
 and no counter-satisfiability was found.
\begin{table*}[htbp]
  \centering
  \caption{Re-proving MPTP 0.2 theorems from non-numerical articles in 2005}
  \begin{tabular}{lrrrrr}
    \toprule
    description&proved&counter-satisfiable&timeout or memory out&total\\\midrule
    E 0.9 &4309&0&8220&12529\\
    SPASS 2.1&3850&0&8679&12529\\
    together &4854&0&7675&12529\\\bottomrule    
  \end{tabular}
\label{Repr1}
\end{table*}

These results have thus to a large extent confirmed the optimistic outlook of the 
first measurement in MPTP 0.1. In later experiments, this ATP 
performance has been steadily going up, see Table~\ref{Repr2} for results from 2007
run with 60s time limit. 
This is a result of better pruning of redundant
axioms in MPTP, and also of ATP development, which obviously was influenced by
the inclusion of MPTP problems in the TPTP library, forming a significant part
of the FOF problems in the CASC competition since 2006. The newer versions of E and SPASS
solved in this increased time limit together 6500 problems, i.e. 52\% of them all. With addition
of Vampire  and its customized Fampire\footnote{Fampire is a combination of Vampire with the FLOTTER~\cite{NonnengartW01} clausifier.} version (which alone solves 51\% of the problems), the combined
success rate went up to 7694 of these problems, i.e. to 61\%. The caveat is that
the methods for dealing with arithmetic are becoming stronger and
stronger in Mizar. The MPTP problem creation
for problems containing arithmetic's is thus currently quite crude, and the ATP success rate on such
problems will likely be significantly lower than on the nonarithmetical ones.
 Fortunately, various approaches have been
recently started to endow theorem provers with better capabilities for
arithmetic~\cite{PW06,KorovinV07,AlthausKW09,BonacinaLM11}.
\begin{table*}[htbp]
  \centering
  \caption{Re-proving MPTP 0.2 theorems from non-numerical articles in 2007}
  \begin{tabular}{lrrrrr}
\toprule
    description&proved&counter-satisfiable&timeout or memory out&total\\\midrule
    E 0.999 &5661&0&6868&12529\\
    SPASS 2.2&5775&0&6754&12529\\
    E+SPASS together &6500&-& - &12529\\
    Vampire 8.1&5110&0&7419&12529\\
    Vampire 9&5330&0&7119&12529\\
    Fampire 9&6411&0&6118&12529\\
    all together &7694&-&-&12529\\\bottomrule
  \end{tabular}
\label{Repr2}
\end{table*}

\subsection{Finding new proofs and the AI aspects}
\label{New}
MPTP 0.2 was also used to try to prove Mizar theorems without looking at their Mizar proofs, i.e., the
choice of premises for each theorem was done automatically, and all previously proved theorems
were eligible. Because giving ATPs thousands of axioms was usually 
hopeless in 2005, the
axiom selection was done by symbol-based machine learning from the
previously available proofs using the SNoW~\cite{Car99} (Sparse Network of Winnows)
system run in the naive Bayes learning mode.
The ability of systems to handle many axioms has improved a lot since 2005: the CASC-LTB category and also the related work on the Sledgehammer~\cite{MengP09} link between ATPs with Isabelle/HOL have 
sparked interest in ATP systems dealing efficiently with large numbers of axioms. 
See~\cite{Urban11-ate} for a brief overview of the large theory methods developed so far.

The results (reported in \cite{Urb07}) are given in Table \ref{PrNew1}. 2408 from the 12529
theorems were proved either by E 0.9 or SPASS 2.1 from the axioms selected by the machine learner, 
the combined success rate of this whole system was thus 19\%. 
\begin{table*}[htbp]
\centering
  \caption{Proving MPTP 0.2 theorems with machine learning support in 2005}
  \begin{tabular}{lrrrrr}
\toprule
    description&proved&counter-satisfiability&timeout or memory out&total\\\midrule
    E 0.9 &2167&0&10362&12529\\
    SPASS 2.1&1543&0&10986&12529\\
    together &2408&0&10121&12529\\\bottomrule
  \end{tabular}
\label{PrNew1}
\end{table*}

This experiment demonstrates a very real and quite unique benefit of large formal mathematical libraries
for conducting novel integration of AI methods.
As the machine learner is trained on previous proofs, it recommends relevant
premises from the large library that (according to the past experience) should be
useful for proving new conjectures. A variety of machine learning methods (neural nets,
Bayes nets, decision trees, nearest neighbor, etc.) can be used for this, and their
performance evaluated in the standard machine learning way, i.e., by looking at the actual
axiom selection done by the human author in the Mizar proof, and comparing it with
the selection suggested by the trained learner. However, what if the machine learner is 
sometimes more clever than the human, and suggests a completely different (and perhaps
better) selection of premises, leading to a different proof? 
In such a case, the standard machine learning evaluation (i.e. comparison of the two sets of premises)
will say that the two sets of premises differ too much, and thus the machine learner has failed.
This is considered acceptable for machine learning, as in general, there is no deeper 
concept of {\it truth} available, there are just training and testing data. 
However in our domain we do have a method for showing
that the trained learner was right (and possibly smarter than the human): we can run an ATP system
on its axiom selection. If a proof is found, it provides a much stronger measure of correctness.
Obviously, this is only true if we know that the translation from Mizar to TPTP was correct, i.e.,
conducting such experiments really requires that we take extra care to ensure that no 
oversimplifications were made in this translation.

In the above mentioned experiment, 329 of the 2408 (i.e. 14\%)
proofs found by ATPs were shorter (used less premises) than the original MML proof. 
An example of such proof shortening is discussed in \cite{Urb07}, showing that the newly found
proof is really valid. Instead of arguing from the first principles (definitions) like in the 
human proof, the
combined inductive-deductive system was smart enough to find a combination of previously 
proved lemmas (properties)
that justify the conjecture more quickly. 

A similar newer evaluation is done on the whole MML in~\cite{AlamaKU12}, comparing the original MML theory graph with the theory graph for the 9141 automatically found proofs. 
An illustrative example from there
is theorem
\texttt{COMSEQ\_3:40}\footnote{\url{http://mizar.cs.ualberta.ca/~mptp/cgi-bin/browserefs.cgi?refs=t40_comseq_3}},
proving the relation between the limit of a complex sequence and its
real and imaginary parts:
\begin{theorem}
Let $(c_n) = (a_n + i b_n)$ be a convergent complex sequence. Then $(a_n)$ and $(b_n)$ converge and
$\lim a_n = Re (\lim c_n)$ and $\lim b_n = Im (\lim c_n)$.  
\end{theorem}
 The convergence of $(a_n)$ and $(b_n)$ was done the same way by the human formalizer and the ATP.
 The human proof of the limit equations proceeds by looking at the definition of a complex
 limit, expanding the definitions, and proving that a and b satisfy the definition of the real limit
(finding a suitable $n$ for a given $\epsilon$).
The
AI/ATP just notices that this kind of groundwork was already done in a ``similar''
case
\texttt{COMSEQ\_3:39}\footnote{\url{http://mizar.cs.ualberta.ca/~mptp/cgi-bin/browserefs.cgi?refs=t39_comseq_3}},
which says that:
\begin{theorem}
 If $(a_n)$ and $(b_n)$ are convergent, then
$\lim c_n = \lim a_n+ i \lim b_n$.
\end{theorem}
 And it also notices the ``similarity'' (algebraic simplification) provided by theorem \texttt{COMPLEX1:28}\footnote{\url{http://mizar.cs.ualberta.ca/~mptp/cgi-bin/browserefs.cgi?refs=t28_complex1}}:
\begin{theorem}
$Re (a + i b) = a \wedge Im (a + i b) = b$
\end{theorem}
 Such (automatically found) manipulations can be  used (if noticed!) to avoid the ``hard thinking'' about the epsilons in the definitions.

\subsection{ATP-based explanation, presentation, and cross-verification of Mizar proofs}
While the whole proofs of Mizar theorems can be quite hard for ATP systems, re-proving
the Mizar atomic justification steps (called {\it Simple
  Justifications} in Mizar; these are the atomic steps discharged by
the very limited and fast Mizar refutational checker~\cite{Wiedijk00}) turns out
to be quite easy for ATPs. The combinations of E and SPASS usually solve more than 95\%
of such problems, and with smarter automated methods for axiom selection 99.8\%
success rate (14 unsolved problems from 6765) was achieved in \cite{US07}. This makes
it practical to use ATPs for explanation and presentation of the (not always easily
understandable) Mizar simple justifications, and to construct larger systems for
independent ATP-based cross-verification of (possibly very long) Mizar proofs.
In \cite{US07} such a cross-verification system is presented, using the GDV~\cite{Sut06}
system which was extended to process Jaskowski-style natural deduction proofs that make 
frequent use of assumptions (suppositions). MPTP was used to translate Mizar proofs
to this format, and GDV together with the E, SPASS, and MaLARea systems were used
to automatically verify the structural correctness of proofs, and 99.8\% of the
proof steps needed for the 252 Mizar problems selected for the MPTP Challenge (see below).
This provides the first practical method for independent verification of Mizar, and
opens the possibility of importing Mizar proofs into other proof assistants.
A web presentation allowing interaction with ATP systems and GDV verification of Mizar
proofs~\cite{UT+07} has been set up,\footnote{\url{http://www.tptp.org/MizarTPTP}} and
an online service~\cite{abs-1109-0616} integrating the ATP functionalities has been 
built.\footnote{\url{http://mws.cs.ru.nl/~mptp/MizAR.html}, and\\ \url{http://mizar.cs.ualberta.ca/~mptp/MizAR.html}}

\subsection{Use of MPTP for ATP challenges and competitions}
The first MPTP problems were included in the TPTP library in 2006, and were already
used for the 2006 CASC competition. In 2006, the MPTP \$100 
Challenges\footnote{\url{http://www.tptp.org/MPTPChallenge/}} were created and announced.
This is a set of 252 related large-theory problems needed for one half (one of two implications) of the
Mizar proof of the general topological Bolzano-Weierstrass theorem~\cite{YELLOW19}.
Unlike the CASC competition, the challenge had an overall time limit (252 * 5 minutes =
21 hours) for solving the problems, allowing complementary techniques like machine learning
from previous solutions to be experimented with transparently within
the overall time limit. The challenge
was won a year later by the leanCoP~\cite{OB03} system, having already revealed
several interesting approaches to ATP in large theories: goal-directed calculi like
connection tableaux (used in leanCoP), model-based axiom selection (used e.g. in SRASS~\cite{SP07}),
and machine learning of axiom relevance (used in MaLARea) from
previous proofs and conjecture characterizations (where \textit{axiom
  relevance} can be defined as the
likelihood that an axiom will be needed for proving a particular conjecture). The MPTP Challenge problems
were again included in TPTP and used for the standard CASC competition in 2007.
In 2008, the CASC-LTB (Large Theory Batch) category appeared for the first time with a similar
setting like the MPTP Challenges, and additional large-theory problems from the SUMO and Cyc~\cite{RRG05}
ontologies. A set of 245 relatively hard Mizar problems was included for this purpose in TPTP,
coming from the most advanced parts of the Mizar library. The problems come in four versions that
contain different numbers of the previously available MML theorems and definitions as axioms.
The largest versions thus contain over 50000 axioms. 
An updated larger version (MPTP2078) of the MPTP Challenge benchmark
was developed in 2011~\cite{abs-1108-3446}, consisting of 2078
interrelated problems in general topology, and making use of precise
dependency analysis of the MML for constructing the easy versions of
the problems. 
The MPTP2078 problems were used in the Mizar category of the 2012
CASC@Turing ATP competition.  This category had similar rules as the
MPTP Challenge, i.e., a large number (400) of related MPTP2078
problems is being solved in an overall time limit. Unlike in the MPTP
Challenge, an additional training set of 1000 MPTP2078 problems
(disjoint from the competition problems) was made available before the
competition together with their Mizar and Vampire proofs. This way,
more expensive training and tuning methods could be developed before
the competition, however (as common in machine-learning competitions)
the problem-solving knowledge extracted from the problems by such
methods has to be sufficiently general to be useful on the disjoint
set of competition problems.

\subsection{Development of larger AI metasystems like  MaLARea and MaLeCoP on MPTP data}
\begin{figure}[thb]
\caption{The basic MaLARea loop.}
\begin{center}
\includegraphics[width=9cm]{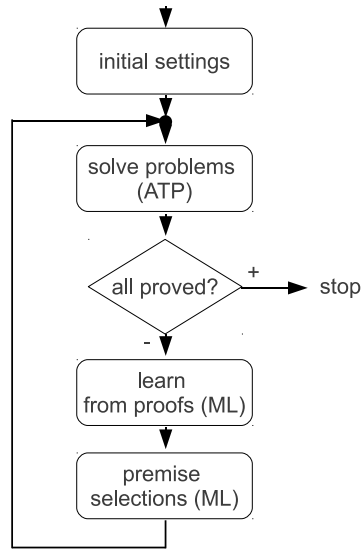}
\end{center}
\label{malarea}
\vspace{-4cm}
\end{figure}

In Section~\ref{New}, it is explained how the notion of mathematical {\it truth}
implemented through ATPs
can improve the evaluation of learning systems working on large semantic knowledge 
bases like translated MML. This is however only one part of the AI fun made possible by
such large libraries being available to ATPs. Another part is that the newly
found proofs can be recycled, and again used for learning in such domains. This
closed loop (see Figure~\ref{malarea}) between using deductive methods to find proofs, and using inductive methods
to learn from the existing proofs and suggest new proof directions, is the main idea
behind the Machine Learner for Automated Reasoning (MaLARea~\cite{Urb07-ESARLT,US+08}) metasystem, which turns out to have so far the best performance on large theory benchmarks like the MPTP Challenge and MPTP2078.
 There are many kinds
of information that such an autonomous metasystem can try to use and learn. The second
version of MaLARea already uses also structural and semantic features of formulas
for their characterization and for improving the axiom selection. 

MaLARea can work with arbitrary ATP backends (E and SPASS by
default), however, the communication between learning and the ATP
systems is \textit{high-level}: The learned relevance is used to try to
solve problems with varied limited numbers of the most relevant
axioms. Successful runs provide additional data for learning (useful
for solving related problems), while unsuccessful runs can yield
countermodels, which can be re-used for semantic
pre-selection and as additional input features for learning.
An advantage of such high-level approach is that it gives a generic
inductive (learning)/deductive (ATP) metasystem to which any ATP can be easily plugged
as a blackbox. Its disadvantage is that it does not attempt to use the
learned knowledge for guiding the ATP search process once the axioms
are selected. 

\begin{figure}[thb]
\caption{The General Architecture used for MaLeCoP}
\begin{center}
\includegraphics[width=8cm]{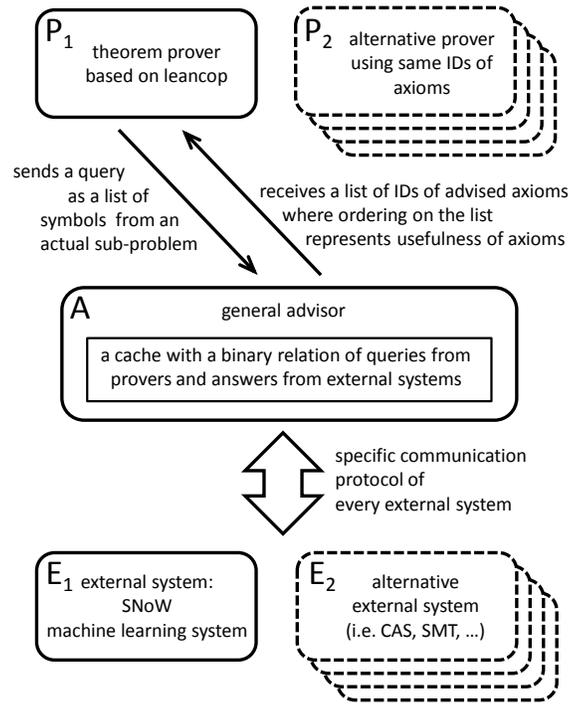}
\end{center}
\label{DesignFig}
\end{figure}

Hence the logical next step done in the Machine Learning Connection Prover (MaLeCoP) prototype~\cite{UrbanVS11}: 
the learned knowledge is used for guiding the proof
search mechanisms in a theorem prover (leanCoP in this case).
MaLeCoP follows a general advising design that is as follows (see also
Figure~\ref{DesignFig}): The theorem prover ({\sf P}) has a
sufficiently fast communication channel to a general advisor ({\sf A}) that
accepts queries (proof state descriptions) and training data
(characterization of the proof state\footnote{instantiated, e.g., as
  the set of literals/symbols on the current branch} together with
solutions\footnote{instantiated, e.g., as the description of clauses
  used at particular proof states} and failures) from the prover,
processes them, and replies to the prover (advising, e.g., which
clauses to choose). The advisor {\sf A} also talks to external (in our case learning) system(s)
({\sf E}). {\sf A} translates the queries and information produced by {\sf P} to the
formalism used by a particular {\sf E}, and translates {\sf E}'s guidance back to
the formalism used by {\sf P}. At suitable time, {\sf A} also hands over the
(suitably transformed) training data to {\sf E}, so that {\sf E} can update its
knowledge of the world on which its advice is based. 

MaLeCoP is a very recent work, which has so far revealed interesting
issues in using detailed smart guidance in large theories. Even though
naive Bayes is a comparatively fast learning and advising algorithm,
in a large theory it turned out to be about 1000 times slower than a
primitive tableaux extension step. So a number of strategies had to be
defined that use the smart guidance only at the crucial points of the
proof search. Even with such limits, the preliminary evaluation done on the MPTP Challenge
already showed an average proof search shortening by a factor of 20 in terms of the number of tableaux inferences.

There are a number of development directions for knowledge-based AI/ATP architectures like MaLARea and MaLeCoP. Extracting lemmas
from proofs and adding them to the set of available premises, creating new interesting
conjectures and defining new useful notions, finding optimal strategies for problem classes,
faster guiding of the internal ATP search, inventing policies
for efficient governing of the overall inductive-deductive loop: all these are interesting AI tasks
that become relevant in this large-theory setting, and that also seem to be relevant
for the task of {\it doing mathematics} (and exact sciences) 
more automatically. A particularly interesting issue is the following.

\subsubsection{Consistency of Knowledge and Its Transfer:}

An important research topic in the emerging AI
approaches to large-theory automated reasoning is the issue of
\textit{consistency of knowledge and its transfer}.  In an unpublished
experiment with MaLARea in 2007 over a set of problems exported by an
early Isabelle/Sledgehammer version, MaLARea quickly solved all of the
problems, even though some of them were supposed to be hard. Larry
Paulson tracked the problem to an (intentional) simplification in
the first-order encoding of Isabelle/HOL types, which typically raises
the overall ATP success rate (after checking in Isabelle the imported
proofs) in comparison with heavier (but sound) type encodings that are also available in Isabelle/Sledgehammer. 
Once the inconsistency originating from the simplification
was found by the guiding AI system, MaLARea focused on fully
exploiting it even in problems where such inconsistency would be
ignored by standard ATP search. 

An opposite phenomenon happened
recently in experiments with a clausal version of MaLARea. The CNF
form introduces a large number of new skolem symbols that make similar
problems and formulas look different after the reduction to clausal form (despite the fact that the skolemization attempts hard to use the same symbol whenever it can), and the AI
guidance based on symbols and terms deteriorates. The same happens
with the AI guidance based on finite counter-models generated by Mace~\cite{McC03-MACE4-TR}
and Paradox~\cite{CS03}. 
The \texttt{clausefilter}
utility from McCune's  Library of Automated Deduction Routines (LADR)~\cite{McC-LADR-URL} is used by MaLARea to evaluate all available formulas in each newly found finite model, resulting in an additional semantic characterization of all formulas that often improves the premise selection.
Disjoint skolem symbols however prevent a
straightforward evaluation of many clauses in models that are found for differently
named skolem functions. Such inability of the AI guidance to obtain and
use the information about the similarity of the clauses results in
about 100 fewer problems solved (700 vs. 800) in the first ten MaLARea
iterations over the MPTP2078 benchmark.

Hence a trade-off: smaller pieces of knowledge (like clauses) allow
better focus, but techniques like skolemization can destroy some
explicit similarities useful for learning. Designing suitable
representations and learning methods on top of the knowledge is
therefore very significant for the final performance, while inconsistent representations can be fatal.
Using CNF and its various alternatives and improvements has been a
topic discussed many times in the ATP community (also for example by
Quaife in his book). Here we note that it is not just the low-level
ATP algorithms that are influenced by such choices of representation,
but the problem extends to and significantly influences also the
performance of high-level heuristic guidance methods in large
theories.

\section{Future QED-like Directions}
\label{Future}
There is large amount of work to be done on practically all the projects mentioned above.
The MPTP translation is by no means optimal (and especially proper encoding of arithmetics 
needs more experiments and work). Import of ATP proofs to Mizar practically does not exist
(there is a basic translator taking Otter proof objects to Mizar, however this is very
distant from the readable proofs in MML). 
With sufficiently strong ATP systems, the  cross-verification of the whole MML could be attempted,
and work on import of such detailed ATP proofs into other proof assistants could be started.
The MPTP handling of second-order constructs is in some sense incomplete, and either a translation
to (finitely axiomatized)  Neumann-Bernays-G{\"o}del (NBG) set theory (used by Quaife), or usage of higher-order ATPs like Satallax~\cite{Brown12} and LEO-II~\cite{BenzmullerPTF08} would be interesting
from this point of view. 

More challenges and interesting presentation tools can be developed,
for example an ATP-enhanced wiki for Mizar is an interesting QED-like project that is now being worked on~\cite{UrbanARG10}. The heuristic and
machine learning methods, and combined AI metasystems, have a very long way to go, some future
directions are mentioned above. This is no longer only about mathematics: all kinds
of more or less formal large knowledge bases are becoming available in other sciences, and
automated reasoning could become one of the strongest methods for general reasoning in sciences
when sufficient amount of formal knowledge exists.
Strong ATP methods for large formal mathematics could also 
provide useful semantic filtering
for larger systems for automatic formalization of mathematical
papers. This is a field that has been so far deemed to be rather
science fiction than a real 
possibility.
In particular, the idea of gradual top-down (semi-)automated formalization of 
mathematics written in books and papers has been so far considered 
outlandish in a way that is quite reminiscent of how the idea of ATP
in large general mathematics was considered outlandish before the first
large-scale experiments in 2003.
The proposed AI solution should be similar for the two problems: guide the vast search space by the knowledge extracted from the vast amount of problems already solved. It is interesting that already within the QED project discussions, Feferman
suggested\footnote{\url{http://mizar.org/qed/mail-archive/volume-2/0003.html}} that large-scale formalization should have a necessary top-down aspect.
Heuristic AI methods
used for knowledge search and machine translation are becoming more
and more mature, and in conjunction with strong ATP methods they could
provide a basis for such large-scale (semi-)automated QED project.

\bibliographystyle{plain}
\bibliography{qed}
\end{document}